\documentclass[runningheads]{llncs}

\usepackage{graphicx}
\usepackage{hyperref}

\begin{document}
\title{ScarGAN: Chained Generative Adversarial Networks to Simulate Pathological Tissue on Cardiovascular MR Scans}
\titlerunning{ScarGAN: Simulating Scars on LGE Scans}
%
\authorrunning{Lau et al.}
\author{Felix Lau\inst{1} \and
Tom Hendriks\inst{1, 2} \and
Jesse Lieman-Sifry\inst{1} \and
Berk Norman\inst{1} \and
Sean Sall\inst{1} \and
Daniel Golden\inst{1}
}
%
%
\institute{Arterys Inc. \\
{51 Federal Street, San Francisco CA 94107, USA} \\
\email{\{felix,tom.hendriks,jesse,berk,sean,dan\}@arterys.com} \\
\and
University of Groningen, University Medical Center Groningen, Department of Cardiology \\
{Hanzeplein 1, (9713 GZ), Groningen, The Netherlands} \\
\email{t.hendriks@umcg.nl} \\
}
\maketitle              
%


\begin{abstract}
Medical images with specific pathologies are scarce, but a large amount of data is usually required for a deep convolutional neural network (DCNN) to achieve good accuracy. We consider the problem of segmenting the left ventricular (LV) myocardium on late gadolinium enhancement (LGE) cardiovascular magnetic resonance (CMR) scans of which only some of the scans have scar tissue. We propose ScarGAN to simulate scar tissue on healthy myocardium using chained generative adversarial networks (GAN). Our novel approach factorizes the simulation process into 3 steps: 1) a mask generator to simulate the shape of the scar tissue; 2) a domain-specific heuristic to produce the initial simulated scar tissue from the simulated shape; 3) a refining generator to add details to the simulated scar tissue. Unlike other approaches that generate samples from scratch, we simulate scar tissue on normal scans resulting in highly realistic samples. We show that experienced radiologists are unable to distinguish between real and simulated scar tissue. Training a U-Net with  additional scans with scar tissue simulated by ScarGAN increases the percentage of scar pixels correctly included in LV myocardium prediction from 75.9\% to 80.5\%.
\end{abstract}


\begin{figure}[t]
    \centering
    \includegraphics[scale=0.40]{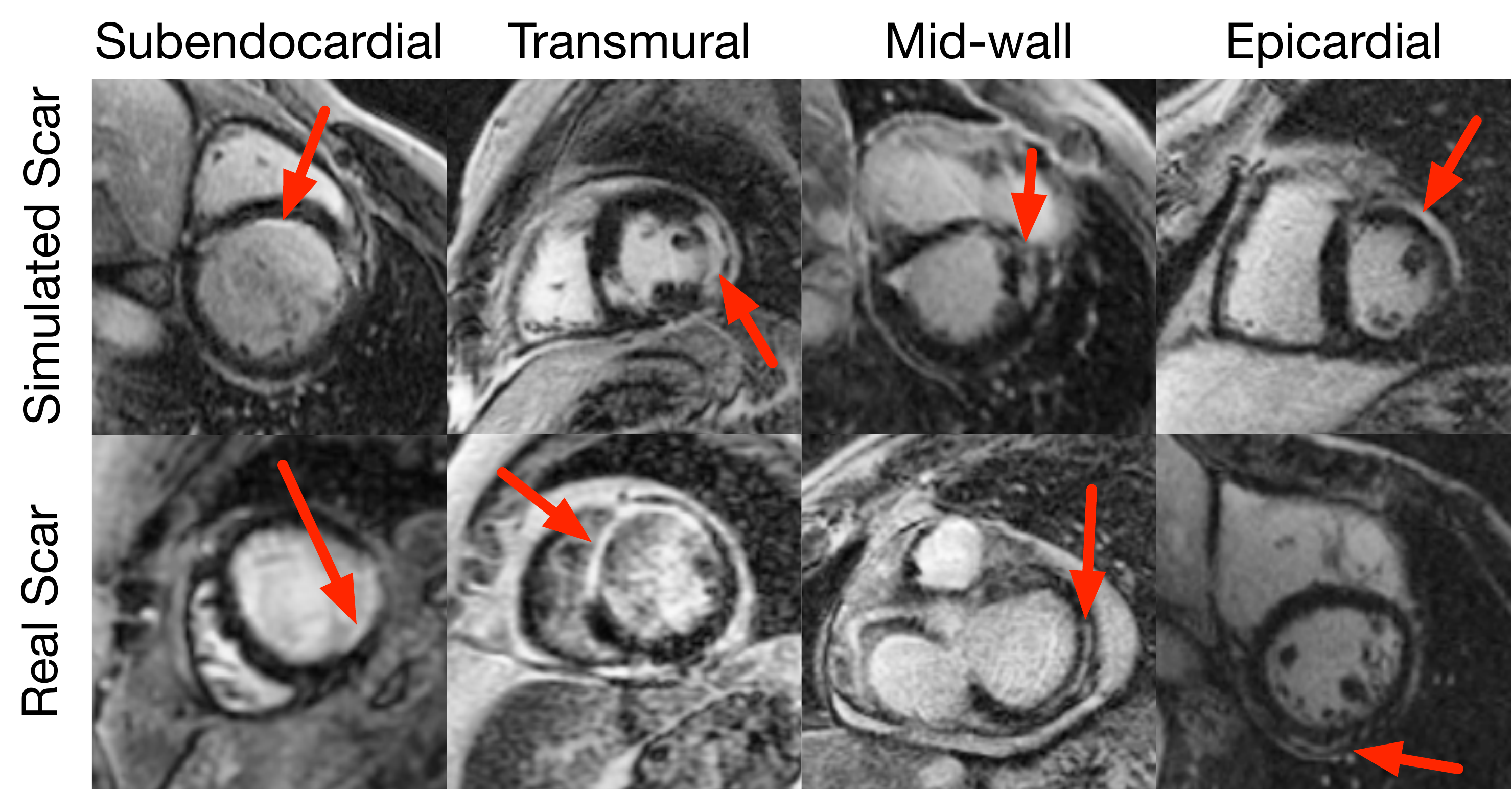}
    \caption{Samples of real and simulated scar tissue on LGE scans categorized by enhancement patterns. Red arrows indicate the locations of real or simulated scar tissue.}
    \label{fig:sim_results}
\end{figure}

\section{Introduction}
Recently, deep learning has shown promising results to automate many tasks in radiology such as skin lesion classification \cite{Esteva2017-ry} and diabetic eye disease detection \cite{Gulshan2016-pc}. The performance of these deep convolutional neural networks (DCNNs) are sometimes on par with clinicians but usually require a large amount of training images and labels. Training a DCNN that performs equally well across different patients is challenging because some pathologies and congenital diseases are rare. Without enough training data for rare pathologies, DCNNs might not perform well on these patients even though they often require the most clinical care.

\subsection{Automated myocardium segmentation LGE scans} Late-gadolinium enhancement (LGE) imaging is an established method to detect myocardial scarring and measure infarct size. Not all LGE scans have visible scar tissue. Contrast accumulates in regions of the myocardium that contain a high proportion of fibrosis (scar tissue) which results in a high signal intensity (hyperenhancement) on the acquired images. Automated myocardium segmentation can be combined with Full-Width-at-Half-Maximum (FWHM) method or n-SD thresholding methods to provide quantitative analysis on LGE scans. \cite{Schulz-Menger2013-xp}

We apply a U-Net segmentation network \cite{Ronneberger2015-hz} to predict segmentation mask of the left myocardium but it does not perform well on patients with scar tissue. The subtle differences between scar tissue and blood pool are extremely challenging for DCNNs and sometimes experts to delineate, especially when the scar tissue is subendocardial. Other challenges in identifying scar tissue include low signal-to-noise ratio, motion blurring and image artifacts. \cite{Karim2016-zu}

\subsection{ScarGAN}
We propose ScarGAN, an approach utilizing chained generative adversarial networks (GANs) to simulate scar tissue in the LV myocardium on LGE scans of healthy patients as data augmentation. Fig. \ref{fig:sim_results} shows examples of simulated scar tissue grouped by their enhancement patterns. Overview of the ScarGAN architecture can be seen in Fig. \ref{fig:scargan}.

ScarGAN has three main components:

\begin{enumerate}
    \item A mask generator $ M(x) $ to generate the shape of the scar tissue given an input segmentation mask of the ventricles;
    \item A domain-specific heuristic to apply the shape of the simulated scar tissue to the image;
    \item A refining generator $ R(x) $ to refine the initial simulation to provide realistic-looking scar tissue.
\end{enumerate}

The main contributions of this work are:

\begin{itemize}
    \item We present ScarGAN to simulate scar tissue in healthy myocardium on LGE CMR scans;
    \item We factorize the simulation process into multiple steps and allow domain-specific heuristics to be added to reduce the difficulty of GAN training;
    \item We present qualitative and quantitative results to demonstrate that scar tissue simulated by ScarGAN is highly realistic and cannot be distinguished from real scar tissue by radiologists;
    \item We demonstrate that simulated scar tissue can improve myocardium segmentation networks without collecting more scans and annotations of a specific pathology.
\end{itemize}


\section{Related Work}

The GAN framework was first proposed by Goodfellow et al. \cite{Goodfellow2014-dw} and consists of 2 networks: a generative network $ G(z) $ that transforms a noise vector $ z $ into realistic samples, and a discriminator network $ D(x) $ that classifies samples as real or fake. The training process of these 2 networks is a minimax game because the objective of $ G(x) $ is to ``fool" $ D(x) $. Redford et al. \cite{Radford2015-fc} propose DCGAN that uses deconvolutional layers in the generator to produce more realistic samples. More recently, pix2pix was proposed by Isola et al. \cite{Isola2016-ks} wherein a fully convolutional U-Net in the generator performs image translation; the generative network receives an image in one domain and outputs the corresponding image of another domain. Both of the GANs in ScarGAN are based on pix2pix.

Previous works have used GANs to refine the results of a simulator (\cite{Shrivastava2016-pf}, \cite{Sixt2016-aa}). SimGAN \cite{Shrivastava2016-pf} proposed by Shrivastava et al, an approach wherein the generator refines the output of a simulator, is structurally similar to our method; however, the "simulator" in ScarGAN is also a GAN and no manual modelling of scar tissue is required.

GANs have been also used in medical imaging for data augmentation. Costa et al. \cite{noauthor_undated-uc} and Zhao et al. \cite{Zhao2017-hc} use pix2pix to generate retinal images from the vessel segmentation mask. Synthetic training samples (such as skin lesions \cite{Baur2018-ka}, liver lesions \cite{Frid-Adar2018-gi} and lung nodules \cite{Chuquicusma2018-ej}) are generated by GAN to increase the size of the training dataset or as educational purposes for radiologists. GANs are also being used to help segmentation networks to work well across different modalities, e.g CycleGAN proposed by Zhu et al. \cite{Zhu2017-lh} to segment brain images in both CT and MRI \cite{Wolterink2017-mo}. Salehinejad et al. \cite{Salehinejad2017-af} use DCGAN to synthesize X-ray chest scans with under-represented diseases and to classify lung diseases. The motivation of the work from Salehinejad et al. is most similar to ours but instead of generating images from scratch, ScarGAN simulates diseases on scans of healthy patients.


\section{ScarGAN}

\begin{figure}[t]
    \centering
    \includegraphics[scale=0.20]{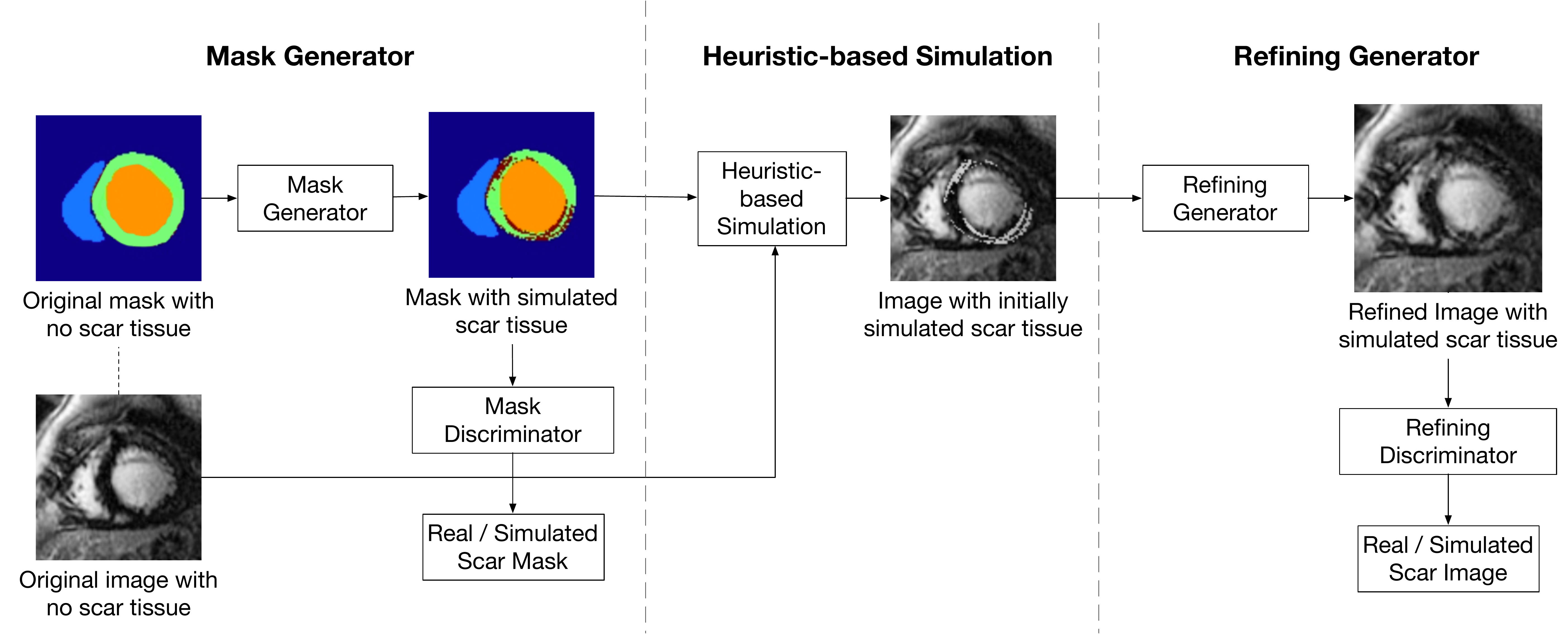}
    \caption{Overview of ScarGAN. A mask generator $ M(x) $ simulates the shape of scar tissue on a segmentation mask; a heuristic-based method provides an initial simulated scar tissue using the simulated shape; a refining generator $ R(x) $ add details of scar tissue to the image.}
    \label{fig:scargan}
\end{figure}

The overall objective of ScarGAN is to learn a function $ f $ that translates an image without scar tissue $ x\textsubscript{no scar} $ to an image with simulated scar tissue $ x\textsubscript{simulated scar} $, i.e $ f: x\textsubscript{no scar} \to x\textsubscript{simulated scar} $ We formulate this problem as an image translation task and thus follow pix2pix. We implement $ f(x) $ as a series of GANs and heuristic-based image processing methods described in detail as follows.

\subsection{Mask Generator $ M(x) $}

In the first stage, ScarGAN uses a mask generator $ M(x) $ to simulate the shape of the scar tissue in the myocardium given an input segmentation mask $ x $ on a slice of a short-axis (SAX) LGE CMR scan. The input of the mask generator is a 192x192px segmentation mask which includes right ventricular endocardium (RV endo), LV myocardium (LV myo) and LV endocardium (LV endo). The output of the mask generator is also a 192x192px segmentation mask which includes simulated scar tissue, RV endo, LV myo and LV endo. Both the input and output include RV endo, LV myo and LV endo to encourage the generator to learn anatomical structures before the discriminator gets too strong which will destabilize training. Unlike pix2pix, the output nonlinearity is softmax instead of tanh because both input and output are segmentation masks instead of images.

The mask generator is a fully convolutional U-Net with 64 initial convolutional filters and skip connections between the corresponding downsampling and upsampling layer blocks. Each of the downsampling or upsampling layer-group consists of a convolutional (or deconvolutional) layer, a batch normalization layer, and a ReLU nonlinearity layer before the downsampling or upsampling operation. Downsampling and upsampling operations are performed via strided convolutions (or deconvolutions.) We add noise to the generator by using dropout layers (p=0.25) after each nonlinearity layer.

The input of the mask discriminator is a 192x192px segmentation mask of RV endo, LV myo and LV endo with real scar tissue or a mask with simulated scar tissue. The discriminator is a relatively shallow network consisting of 4 layer blocks, each contains a convolutional layer, a batch normalization layer, and a leaky ReLU nonlinearity layer ($ \alpha = 0.2 $). The discriminator also downsamples by using strides in its convolutional layers. Unlike pix2pix, we do not follow PatchGAN in which the discriminator classifies patches of the images, and instead classify the whole image as real or simulated.

Following LSGAN \cite{Mao2016-mi}, we use squared error as the main loss function. Thus the discriminator has no nonlinearity in its last layer. To regularize the generator, we add multi-class cross-entropy between the input and output segmentation masks to encourage the network to produce reasonable masks. We train the mask discriminator for 2 gradient steps for every gradient step performed on the mask generator. The loss of the mask generator is as follows:

$$
L_{M} = (1 - D_M(M(x\textsubscript{no scar})))^2 + \alpha xent(M(x\textsubscript{no scar}), x\textsubscript{no scar})
$$

where $ D_M $ is the mask discriminator, $ \alpha $ is a hyperparameter which controls the strength of regularization where we set $ \alpha = 1.0 $, $ xent $ is the per-pixel multiclass cross-entropy, $ x\textsubscript{no scar} $ is a segmentation mask with no scar tissue and $ x\textsubscript{real} $ is a segmentation mask with real scar tissue.

To prevent mode collapse, half of the simulated masks are stored in a buffer for ``experience replay" (\cite{Pfau2016-cg}, \cite{Shrivastava2016-pf}). Half of the training batches for the discriminator are randomly drawn from this buffer. The previously simulated samples stabilize training for both the discriminator and generator and prevent the generator from exploiting the discriminator by generating scar tissue of one specific shape which the discriminator has ``forgotten".

\subsection{Heuristic-based simulation} \label{heuristic-based-simulation}

In the second stage, we apply the shape of the simulated scar tissue from the mask to the image using a heuristic-based method, leveraging the domain-specific knowledge that scar tissue is hyperintense and has a similar signal intensity to the LV blood pool. We replace the corresponding pixels within generated scar tissue with the 10th percentile intensity of the LV endo pixels. However this causes the intensities within the scar tissue to become uniform, which is not characteristic of real scar tissue. 

Although this approach does not produce photorealistic simulation of the scar tissue, it provides a good starting point for another GAN to refine the initial simulation. The result of this initial heuristic-based simulation can be seen in Fig. \ref{fig:scargan}.

\subsection{Refining generator $ R(x) $}

In the final stage, ScarGAN uses a refining network $ R(x) $ to add details to the initial simulation from the heuristic-based method described above. This stage is inspired by SimGAN \cite{Shrivastava2016-pf} but $ R(x) $ in ScarGAN refines results from another GAN instead of a simulator. The input to the refining generator is a 192x192px image with a heuristic-based simulated scar tissue described in Section \ref{heuristic-based-simulation} and its output is a 192x192px refined image with simulated scar tissue. The network architectures of the refining generator and refining discriminator are the same as those described in the mask generator section. We do not need to increase the capacity of the generator given the initial simulation provide a good starting point.

Similar to the mask generator, we follow LSGAN and use squared error as the main loss function. To regularize the generator, we use absolute error between the input and the refined image, which encourages the generator to modify small regions of the image. We train the discriminator for 3 gradient steps for every 1 gradient step performed on the refining generator. The loss of the refining generator is as follows:

$$
L_{R} = (1 - D_R(R(x\textsubscript{scar heuristic})))^2 + \alpha (R(x\textsubscript{scar heuristic}) - x\textsubscript{scar heuristic})
$$

where $ D_R $ is the refining discriminator, $ \alpha $ is a hyperparameter which controls the strength of regularization, and $ x\textsubscript{scar heuristics} $ is a SAX slice with scar tissue simulated by the heuristic method.

We note that $ \alpha $ is a very important hyperparameter as it controls how similar the refined image should be with the input image. If $ \alpha $ is too high, the refining network will be unable to change much of the image; if it is too low, the network might produce artifacts outside the myocardium to exploit the discriminator. We empirically find that setting $ \alpha = 10 $ provides a good balance.

We also use an experience replay buffer to store past simulated samples but we find that mode collapse is not an issue in $ R(x) $ because the image is heavily conditioned by the shape of scar tissue generated by $ M(x) $.

\section{Results and Evaluation}

\subsection{Dataset}

We evaluate our approach using a dataset with 159 LGE SAX CMR scans of which 69 have visible scar tissue on at least one of the SAX slices. We note that our labelled dataset is curated to have a higher proportion of scans with scar tissue (43.4\%) than in our unlabelled dataset ($ \approx25\% $). Our dataset consists of scans acquired by multiple types of scanners across multiple regions and countries. 

Ground truth segmentation masks are collected from 3 physicians with extensive experience analyzing CMR scans. RV endo, LV myo and LV endo segmentation masks are collected by drawing splines around the ventricles. Ground truth scar masks are collected using the FWHM method in accordance with the Society of Cardiovascular Magnetic Resonance imaging (SCMR) guidelines \cite{Schulz-Menger2013-xp}. One small region-of-interest (ROI) is drawn on part of the scar tissue and the FWHM method is applied to derive the full ground truth scar mask. No ROI is drawn if there is no visible scar tissue. Even though FWHM is the recommended method, it achieve relatively low precision of the actual scar tissue \cite{Zhang2016-mo}. In cases where the presence of scar tissue is ambiguous, physicians are able to view LGE scans of other acquisition angles (such as the long-axis) if available, but no segmentation is collected in views other than SAX.

\subsection{Generating a Dataset with Diverse Simulated Scar Tissue using ScarGAN}

\begin{figure}
    \centering
    \includegraphics[scale=0.38]{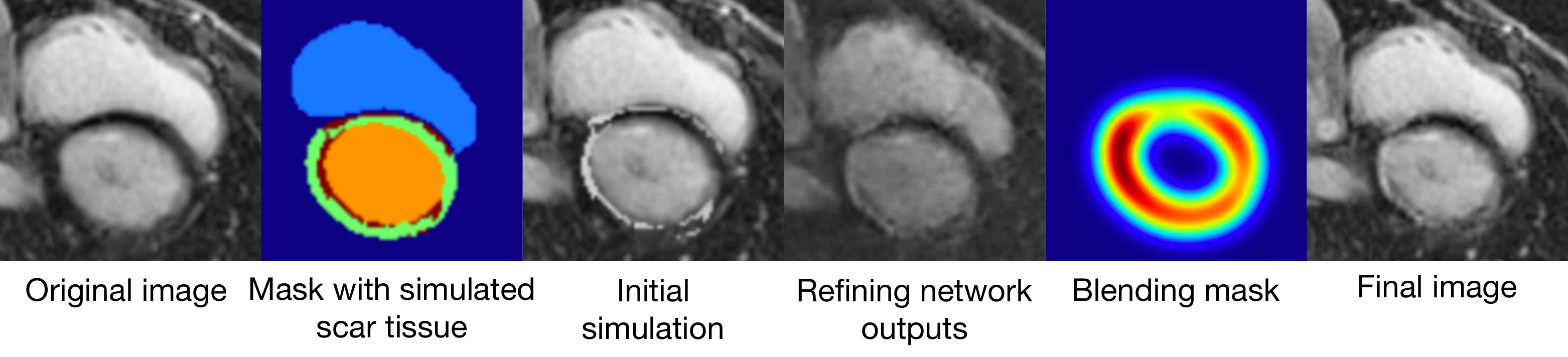}
    \caption{ScarGAN simulation pipeline. From left to right: original image with no scar tissue; mask with RV endo (light blue), LV myo (green), and LV endo (orange) and simulated scar tissue (red); heuristic-based simulation; output from $ R(x) $; blending mask; final image with simulated scar tissue.}
    \label{fig:sim_process}
\end{figure}

In this section, we describe in detail the pipeline we use to simulate scar tissue on healthy myocardium in order to augment the LGE dataset. This dataset is used to train a segmentation network to segment RV endo, LV epicardium (LV epi) and LV endo. Although it is possible to train a network to directly segment scar tissue, analyzing LGE scans using scar tissue segmentation is not part of the recommended guidelines. \cite{Schulz-Menger2013-xp}

Despite using an experience replay buffer, we still sometimes observe mode collapse in $ M(x) $: we monitor the simulated scar tissue throughout the training process and notice that $ M(x) $ sometimes predicts scar tissue of similar shapes in most of the training set. To obtain diverse scar tissue, we try to condition the input by a noise vector or inject noise using dropout at inference time. However neither of these methods yield scar tissue with significantly different shapes. We note that this phenomenon is consistent with the findings in pix2pix. \cite{Isola2016-ks}

Instead, we snapshot the weights of $ M(x) $ for every 10,000 training steps. We pick 5 of these snapshots by visually inspecting the shape of some of the simulated scar tissue. These weight snapshots are selected if simulated scar tissue is of relatively diverse shapes within the training set and across the snapshots.

As a final post-processing step, we create a blending mask to combine the refined image from $ R(x) $ and the original image. This step removes any artifacts created by $ R(x) $ outside the myocardium. We initialize the blending mask as the myocardium mask and then apply Gaussian blur with a kernel size of 5px because the boundary between LV blood pool and LV myocardium is not clear-cut. The final image is created by computing a per-pixel weighted average between the refined image from $ R(x) $ and the original image.

\subsection{Quantitative Evaluation of Simulated Scar Tissue}


\begin{figure}
    \centering
    \includegraphics[scale=0.45]{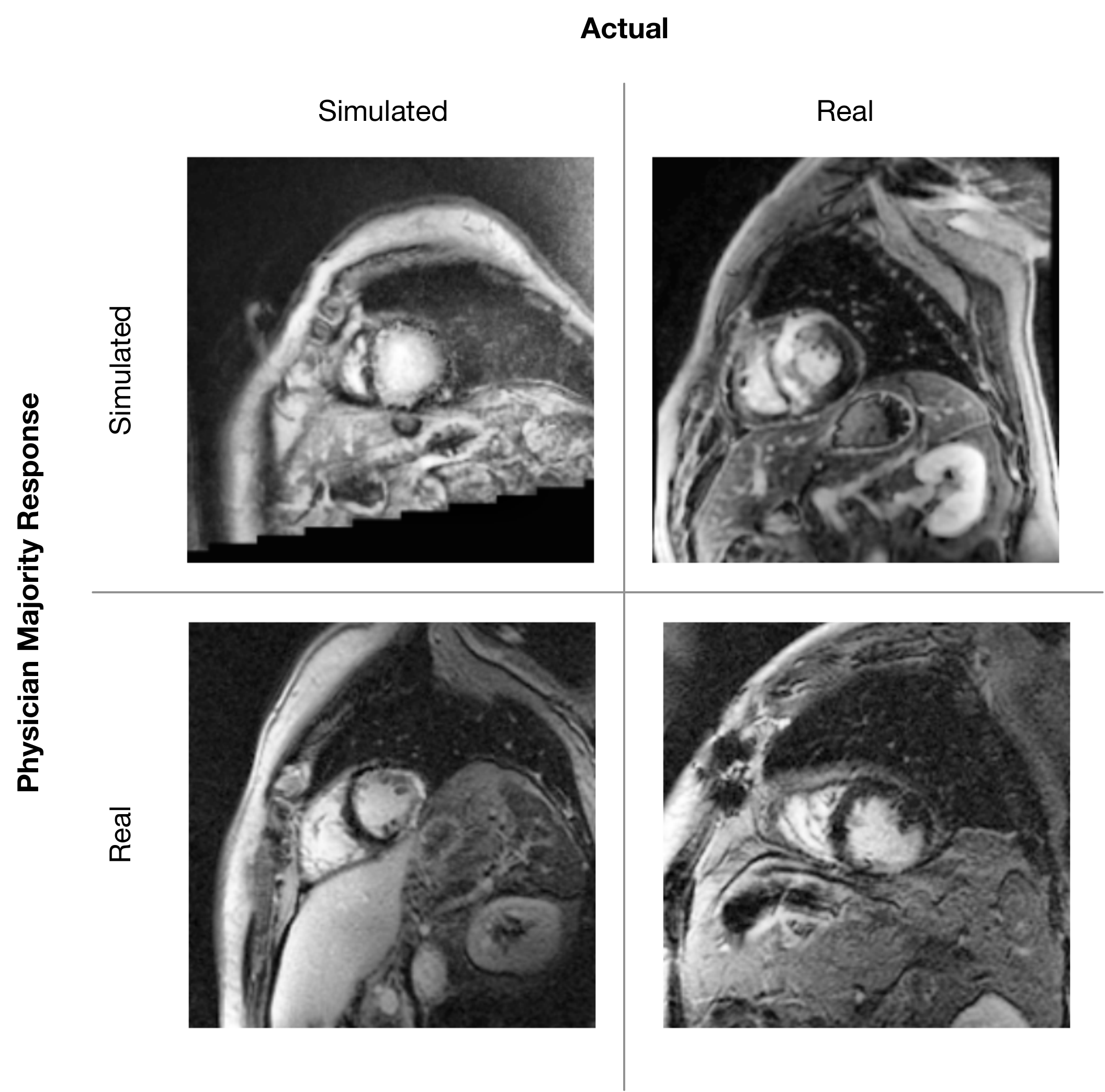}
    \caption{Samples of simulated and real scar tissue grouped by the majority consensus responses from the physicians.}
    \label{fig:visual_turing_test}
\end{figure}

To quantitatively evaluate the realism of the scar tissue simulated by ScarGAN, we ask 3 physicians, including 2 radiologists with more than 10 years of experience, to classify whether the scar tissue on 30 LGE SAX slices are simulated or not. 15 slices have real scar tissue and 15 have scar tissue simulated by ScarGAN. These slices are shown in random order and are randomly drawn from a held-out set. We do not impose any time limit for the physicians to classify the images. The classification accuracies of the physicians are 60\% (p=0.388), 47\% (p=0.804) and 50\% (p=0.607). The majority consensus has an accuracy of 53\% (p=0.791). The results demonstrate that experienced physicians are unable to reliably distinguish simulated scar tissue from real scar tissue and that scar tissue simulated by ScarGAN are highly realistic. Fig. \ref{fig:visual_turing_test} shows samples of simulated and real scar tissue grouped by the majority consensus responses.

\subsection{Segmenting LGE Scans}

\begin{figure}
    \centering
    \includegraphics[scale=0.56]{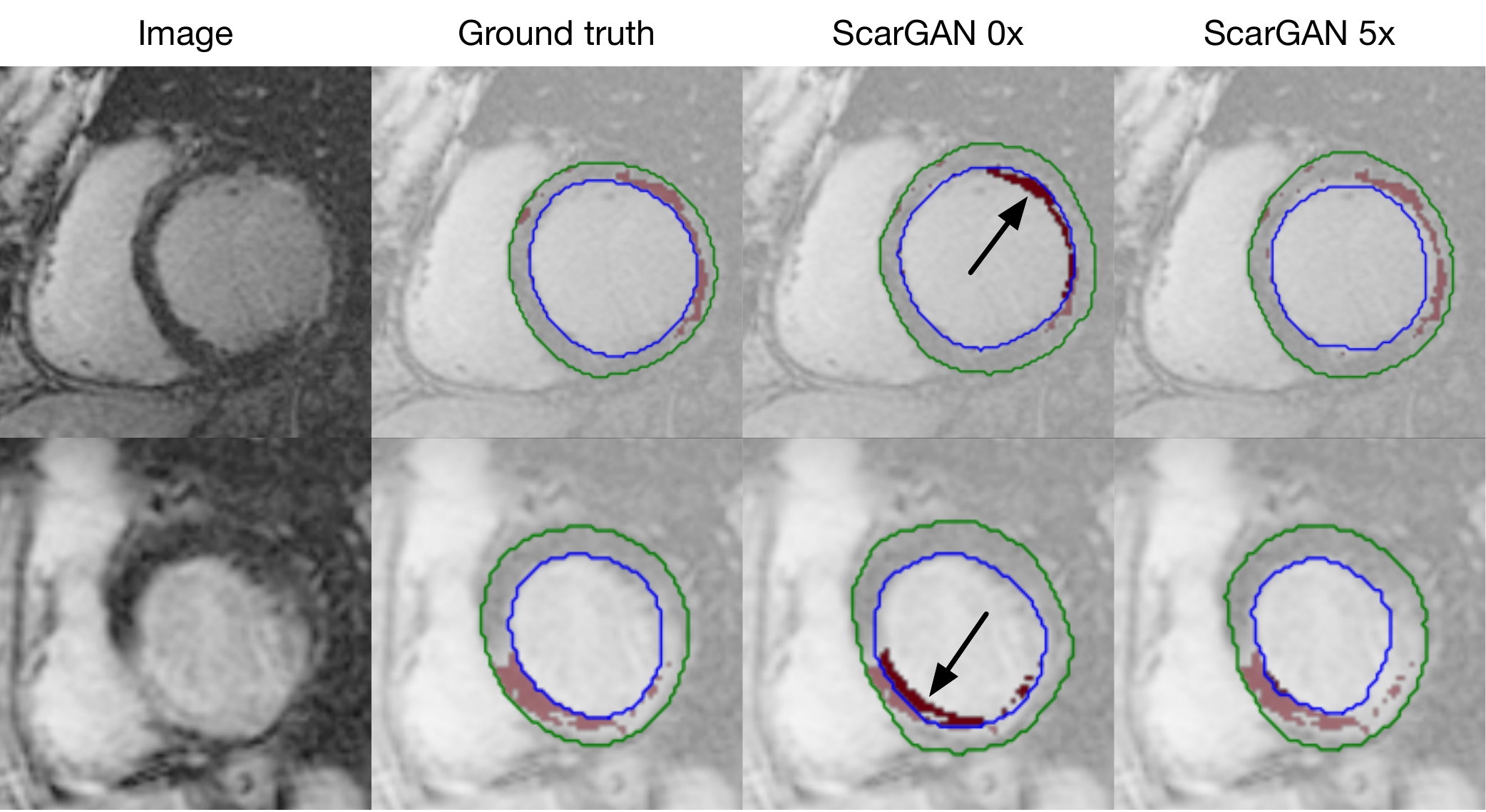}
    \caption{Comparison of predicted and ground truth contours on the LGE SAX images from 2 unique patients. Light red mask is scar tissue correctly included in myocardium; dark red mask is scar tissue erroneously included in blood pool (also indicated by black arrows). The green and blue lines show the predicted LV epi and LV endo contours, respectively.}
    \label{fig:pred_contours}
\end{figure}

We evaluate the effectiveness of ScarGAN as a data augmentation technique by adding scans with simulated scar tissue in the training dataset of a segmentation network that segments RV endo, LV endo and LV epi. We note that LV myo can be derived by subtracting LV endo from LV epi at inference time.

The segmentation network is a U-Net-based DeepVentricle \cite{Lieman-Sifry2017-cm} with 96 initial filters, 4 downsampling layers, 3 convolutional layers before each downsampling or upsampling operations, and skip connection between the corresponding downsampling and upsampling layers. This segmentation network is first trained on 1143 steady-state free precession (SSFP) SAX CMR scans with per-pixel cross-entropy as the objective function and optimized using Adam \cite{Kingma2014-kl} with a learning rate of $ 1e^{-4} $. The network is then fine-tuned on the 159 LGE CMR scans dataset described above. We note that scar tissue is not visible in standard SSFP scans as no contrast agents have been administered. Thus the network must learn all knowledge about scar tissue from the LGE dataset. We apply traditional data augmentation to all networks such as translation, scale, rotation and elastic deformation.

We first naively fine-tune U-Net on the LGE dataset but we find that it is unable to discriminate between blood pool and scar tissue, which have similar intensities. We make two changes to the LGE-specific segmentation network to address this failure mode:

\begin{enumerate}
    \item Other than the three ventricular structures, the LGE segmentation network also predicts the ground truth scar tissue mask derived from FWHM or the simulated scar mask as an auxiliary task.
    \item We modify the loss function such that a higher weighting is assigned to the pixels within scar tissue. We note that this weighting is an important hyperparameter -- if it is too high, the network will overestimate the proportion of scar tissue by erroneously including blood pool as part of the myocardium; if it is too low, the proportion of scar tissue will be underestimated by erroneously missing scar tissue in the myocardium prediction.
\end{enumerate}

We evaluate the model on the percentage of scar tissue pixels erroneously included in LV endo and the percentage of scar tissue pixels correctly included in LV myo in each scan. We perform 4-fold cross-validation on the segmentation network and ScarGAN networks. The dataset is split using anonymized patient IDs such that all scans of one patient are assigned to a single fold. We keep the validation set intact and no scans with simulated scar tissue are added to the validation set.

As shown in Table \ref{tab:eval_results}, we train the segmentation models on different subsets of the dataset: 1) ScarGAN 0x: only scans with scar tissue, 2) ScarGAN 0x+: all scans with and without scar tissue, 3) ScarGAN $ kx $: scans with real and simulated scar tissue where $ k $ is the number of $ M(x) $ weight snapshots we used to simulate scar tissue. We notice that adding scans without scar tissue is detrimental to the network’s ability to distinguish scar tissue. In contrast, adding simulated scar tissue reduces the average percentage of scar tissue pixels erroneously included in LV endo from 10.66\% to 7.55\%, and increases the average percentage of scar tissue pixels correctly included in LV myo from 75.9\% to 80.5\%. The mean LV endo and LV epi contour dice coefficients are 0.869 and 0.906. Fig. \ref{fig:pred_contours} shows predicted and ground truth LV endo and LV epi contours in the test set. This indicates that it is possible to improve model performance on patients with pathologies by collecting scans without any pathologies and using ScarGAN to simulated those pathologies. 

\begin{table}[]
    \centering
    \resizebox{\textwidth}{!}{
        \begin{tabular}{ | c | c | c | c | }
            \hline
                \textbf{Dataset subset} & \textbf{Training data} & \textbf{\% of scar} & \textbf{\% of scar} \\
                &  (number of scans with real scars & \textbf{in LV myo} & \textbf{in LV endo} \\
                &  / no scars / simulated scars) & & \\
            \hline
                ScarGAN 0x & 69 / 0 / 0 & 75.9 (2.1) & 10.7 (1.8) \\
            \hline
                ScarGAN 0x+ & 69 / 90 / 0 & 71.8 (2.4) & 14.1 (2.2) \\
            \hline
                ScarGAN 1x & 69 / 0 / 90 & 79.7 (2.2) & 7.6 (1.3) \\
            \hline
                ScarGAN 3x & 69 / 0 / 270 & 79.2 (2.2) & 8.4 (1.6) \\
            \hline
                ScarGAN 5x & 69 / 0 / 450 & \textbf{80.5 (2.0)} & \textbf{7.5 (1.4)} \\
            \hline
        \end{tabular}
    }
    \newline
    \caption{Evaluation metrics of model trained on LGE scans with and without simulated scars.}
    \label{tab:eval_results}
\end{table}

\section{Conclusions and Future Works}

We propose ScarGAN, a framework that can simulate scar tissue on LGE scans of healthy patients to reduce the need to collect scans from patients with rare pathologies. Unlike existing generative approaches, we factorize the simulation process into multiple steps to reduce the difficulty of training generative networks, and allow domain-specific knowledge to be included in the simulation process. We find that scans with simulated scar tissue cannot be distinguished by physicians and they can improve segmentation performance without additional data and annotation collection.

In the future, we can evaluate ScarGAN with other pathologies and on other tasks such as classification. We can also train both the mask generator and refining generator end-to-end. Further improvements in GAN techniques can be incorporated including the use of Wasserstein distance \cite{Gulrajani2017-gu} and minibatch discrimination layer \cite{Karras2017-bu}.

%
%
%
\bibliographystyle{splncs04}
\bibliography{scargan}
\end{document}